\documentclass[preprint,12pt]{elsarticle}
\usepackage{float}
\usepackage{graphicx}
\usepackage{makecell}
\usepackage[dvipsnames,svgnames,x11names]{xcolor}
\usepackage{makecell}
\usepackage{algorithm}
\usepackage{algpseudocode}

\usepackage[utf8]{inputenc}
\usepackage[margin=1in]{geometry}
\usepackage{booktabs}
\usepackage{siunitx} 
\usepackage{multirow}
\usepackage{tabu}
\usepackage[labelfont = bf, skip=5 pt, font= small]{caption}
\usepackage{subcaption}
\usepackage{amsmath,amssymb}
\usepackage{stmaryrd}
\usepackage{graphicx}
\usepackage{fancyhdr}
\usepackage{float}
\usepackage{mathrsfs}
\journal{arXiv}

\begin{document} 


\title{Adaptive Interface-PINNs (AdaI-PINNs): An Efficient Physics-informed Neural Networks Framework  for Interface Problems}
\author[add1]{\small Sumanta Roy}
\author[add1]{Chandrasekhar Annavarapu\corref{cor1}}
\cortext[cor1]{Corresponding author}
\ead{annavarapuc@civil.iitm.ac.in}
\author[add2]{Pratanu Roy}
\author[add3]{Antareep Kumar Sarma}
\address[add1]{Department of Civil Engineering, Indian Institute of Technology, \\
         Madras, Chennai 600036, Tamil Nadu, India}
\address[add2]{Atmospheric, Earth and Energy Division, Lawrence Livermore National Laboratory, \\ Livermore, California 94551, USA}
\address[add3]{Institute of Civil Engineering,
         École Polytechnique Fédérale de Lausanne, \\ Station 18-1015, Lausanne, Switzerland}


\date{}
\begin{frontmatter}
\begin{abstract}
\sloppy
We present an efficient physics-informed neural networks (PINNs) framework, termed Adaptive Interface-PINNs (AdaI-PINNs), to improve the modeling of interface problems with discontinuous coefficients and/or interfacial jumps. This framework is an enhanced version of its predecessor, Interface PINNs or I-PINNs (Sarma et al.; https://dx.doi.org/10.2139/ssrn.4766623), which involves domain decomposition and assignment of different predefined activation functions to the neural networks in each subdomain across a sharp interface, while keeping all other parameters of the neural networks identical. In AdaI-PINNs, the activation functions vary solely in their slopes, which are trained along with the other parameters of the neural networks. This makes the AdaI-PINNs framework fully automated without requiring preset activation functions. Comparative studies on one-dimensional, two-dimensional, and three-dimensional benchmark elliptic interface problems reveal that AdaI-PINNs outperform I-PINNs, reducing computational costs by 2-6 times while producing similar or better accuracy. 
\end{abstract}


\end{frontmatter}

\section{Introduction}
\label{sec:introduction}
In recent years, Physics-informed neural networks (PINNs) has become increasingly popular as a numerical method for solving partial differential equations (PDEs). Unlike traditional data-driven methods that rely on large training datasets, PINNs require minimal training data because the loss functional in PINNs is constructed from the residuals of the governing equations, boundary and/or initial conditions~\cite{lagaris1998artificial,raissi}. Additionally, the meshless character of PINNs makes this approach potentially suitable for solving a wide range of complex engineering problems (e.g., see~\cite{cuomo2022scientific, lawal2022physics, roy2024exact}).

Despite its widespread use, the application of PINNs to interface problems is still relatively uncommon. Interfaces can cause weak or strong discontinuities in field variables, posing challenges for numerical methods that depend on underlying grids to construct approximate solutions. Grid-based methods often require conforming meshes or intrusive modifications to data structures to appropriately treat discontinuities in the field variables~\cite{annavaraputhesis,hansbo2005nitsche}. The meshless nature of PINNs makes it inherently suited for such problems. Over the past few years, notable attempts have been made to solve interface problems with PINNs~\cite{zhang2023physics, zhang2022physics,aliakbari2023ensemble}. However, all these methods involve significant modifications to the underlying PINNs architecture. Extended PINNs (XPINNs) and conservative PINNs (cPINNs) frameworks of Jagtap et al.~\cite{jagtap2020conservative,jagtap2021extended} provided breakthrough enhancements to the traditional PINNs framework, although still in the context of homogeneous materials, by introducing domain decomposition and prescribing flux continuity constraints at fictional interfaces. These ideas were subsequently incorporated to model fluid flow and heat conduction with discontinuous material coefficients in several follow-up studies~\cite{zhang2022multi,bandai2022forward}. However, in these methods, the neural networks (NNs) used in each fictitious subdomain were different; therefore, the methods require significantly more trainable parameters than conventional PINNs.

Recently, Sarma et al.~\cite{sarma2024ipinns, sarma2023variational} developed a novel physics-informed neural networks framework for modeling interface problems called Interface PINNs (I-PINNs). I-PINNs uses different neural networks for any two subdomains separated by a sharp interface, with the neural networks differing only in their activation functions (AFs) while sharing the same set of parameters (weights and biases). Through benchmark elliptic interface problems, it was demonstrated that I-PINNs outperformed both conventional PINNs and other domain decomposition PINNs frameworks such as multi-domain PINN (M-PINN)~\cite{zhang2022multi} and X-PINNs~\cite{jagtap2021extended} in both accuracy and cost. 

The key idea of the I-PINNs framework is to utilize different AFs for different subdomains. However, as the number of subdomains increases, identifying these AFs can sometimes be challenging. While using the same AFs across alternating subdomains could mitigate this issue, doing so adversely affects convergence rate of I-PINNs. The objective of this study is to address this limitation by automating the choice of AFs for each subdomain using the concept of adaptive activation functions (AAFs)~\cite{jagtap2020adaptive}. The parameters for the AAFs are trained alongside the usual weights and biases to provide an optimal choice of AFs for a given problem. 

The rest of the paper is structured as follows. Section~\ref{sec_model} provides the governing equations for a model elliptic interface problem. Section~\ref{sec:methodology} provides an overview of the I-PINNs and adaptive I-PINNs frameworks. Section~\ref{sec_experiments} presents the results from numerical experiments on one-dimensional (1D), two-dimensional (2D), and three-dimensional (3D) elliptic interface problems where the I-PINNs and the adaptive I-PINNs (AdaI-PINNs) frameworks are compared. Finally, in Section~\ref{sec:conc}, a summary and an outlook of this work are provided.

\section{Model Problem}\label{sec_model}
\begin{figure}[!hbt]
\begin{centering}
\includegraphics[width=0.3\textwidth]{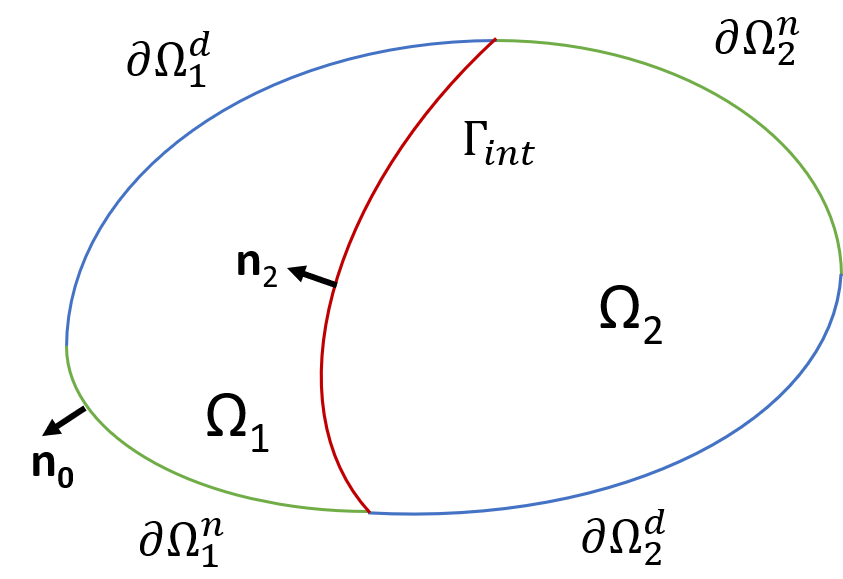}
\par\end{centering}
\caption{Schematic of the problem domain with two regions $\Omega_1$ and $\Omega_2$ separated by an interface $\Gamma_{\text{int}}$.}\label{fig:ProbSetUp}
\end{figure}

We consider a typical Poisson's problem with discontinuous coefficients in a domain $\Omega$, with a boundary $\partial \Omega$, partitioned into two non-overlapping subdomains, $\Omega_1$, and $\Omega_2$ by a sharp interface $\Gamma_\text{int}$ as shown in Figure~\ref{fig:ProbSetUp}. The boundary $\partial\Omega$ has an outward pointing normal $\mathbf{n_0}$ and is partitioned into disjoint sets $\partial\Omega_1$ and $\partial\Omega_2$ such that $\partial\Omega = \overline{\partial\Omega_1\cup\partial\Omega_2}$. 
The governing equations in each domain $\Omega_\text{m}$ (for m = 1,2) are 
\begin{align}
  \begin{split}
    \nabla\cdot(\kappa_\text{m}\nabla u_\text{m}) &= -f_\text{m} \qquad \text{in}~~\Omega_\text{m},\\
    u_\text{m} &= \Lambda^{\text{d}}_\text{m}\qquad \text{on}~~\partial\Omega^\text{d}_\text{m},\\
    \kappa_\text{m}\nabla u_\text{m}\cdot\mathbf{n_0} &= \Lambda^\text{n}_\text{m}\qquad \text{on}~~\partial\Omega^\text{n}_\text{m},
  \end{split}
  \label{eq:Model_int_eqn}
\end{align}
where $\nabla$ is the gradient operator, and $\kappa_\text{m} > 0$, and $f_\text{m}$ are given scalar fields that can be discontinuous across the interface. The disjoint sets $\partial\Omega^\text{d}_\text{m}$ and $\partial\Omega^\text{n}_\text{m}$ represent the Dirichlet and Neumann portions of the boundary such that $\partial\Omega_\text{m} = \overline{\partial\Omega^\text{d}_\text{m}\cup\partial\Omega^\text{n}_\text{m}\cup\Gamma_\text{int}}$. The prescribed values of the boundary data for the primary variable and the flux are $\Lambda^\text{d}_\text{m}$ and $\Lambda^\text{n}_\text{m}$, respectively. The following jump conditions are provided at the interface $\Gamma_\text{int}$ 

\begin{align}
    \begin{split}
      \llbracket u\rrbracket &=p \qquad \text{on}~~\Gamma_\text{int},\\
      \llbracket \kappa \nabla{u}\rrbracket \cdot{\mathbf{n_2}} &= q \qquad \text{on}~~\Gamma_\text{int},
      \label{eq:Model_int_cond}
    \end{split}
\end{align}
where $\mathbf{n_2}$ is the normal to the interface and points away from $\Omega_2$. The bracket operator $\llbracket \odot \rrbracket = \odot_2 - \odot_1$ represents a jump in the quantity $\odot$, and $p$ and $q$ are given scalar fields at the interface.

\section{Methodology}\label{sec:methodology}
\subsection{Vanilla I-PINNs}\label{sec:vanilla_ipinns}
I-PINNs utilizes deep neural networks to construct an approximation  ${u}^\theta_\text{m}(\mathbf{x}_\text{m}, \boldsymbol{\theta})$ to the field variable $u_\text{m}(\mathbf{x}_\text{m})$ in each subdomain $\Omega_\text{m}$, that satisfies the governing equations and boundary conditions specified in Eqs.~\eqref{eq:Model_int_eqn}--\eqref{eq:Model_int_cond}. A deep neural network approximation, ${u}^\theta_\text{m}(\mathbf{x}, \boldsymbol{\theta})$, with a depth of \( D \), which includes an input layer, \( D-1 \) hidden layers, and an output layer, can be expressed as 
\begin{equation}
    {u}^\theta_\text{m}(\mathbf{x}_\text{m}, \boldsymbol{\theta}) = f_\text{D} \circ f_\text{D-1} \circ \ldots \circ f_1({\mathbf{x}_\text{m}}, \boldsymbol{\theta}). 
\end{equation}
where, the symbol \( \circ \) denotes the composition of functions, such that $$f_\text{D}\circ f_\text{D-1} = f_\text{D}(f_\text{D-1}(\mathbf{x}_\text{D-1}, \boldsymbol{\theta}_\text{D-1}))$$ is a recursive operation. At any layer s, in the m-th subdomain, the function $f_\text{s}$ is defined as $f_\text{s}({\mathbf{x}_\text{m}}_\text{s}, \boldsymbol{\theta}_\text{s}) = \sigma_\text{m}(\mathbf{w}_\text{s}^T {\mathbf{x}_\text{m}}_\text{s} + \mathbf{b}_\text{s})$. In this equation, \(  {\mathbf{x}_\text{m}}_\text{s} \) denotes the input vector to the s-th layer in the m-th subdomain, and \( \boldsymbol{\theta}_\text{s} \) represents the parameters of that layer, which includes the weight matrix \( \mathbf{w}_\text{s} \) and the bias vector \( \mathbf{b}_\text{s} \). The function \( \sigma_\text{m}(\odot) \) is the AF applied to the output of each neuron in each subdomain $\Omega_\text{m}$, introducing nonlinear characteristics into the approximation. It is to be noted that the core philosophy of I-PINNs is that the parameter set $\boldsymbol{\theta}$ is the same for each subdomain m, but the AF \( \sigma_\text{m} \) is different. 

For illustration, the loss functional $\zeta(\boldsymbol{\theta})$ for solving the Eqs.~\eqref{eq:Model_int_eqn}-\eqref{eq:Model_int_cond} with a single interface and two subdomains is defined below
\begin{align}
      \zeta(\boldsymbol{\theta}) &= \text{MSE}_\text{eq} + \alpha_\text{bc}^\text{d}\text{MSE}_\text{bc}^\text{d}+ \alpha_\text{bc}^\text{n}\text{MSE}_\text{bc}^\text{n} + \alpha_\text{int}\text{MSE}_\text{ic}^\text{d} + \alpha_\text{int}\text{MSE}_\text{ic}^\text{n}
    \label{eq:loss_function_ipinns}
\end{align}
where $\alpha_\text{bc}^\text{d}, \alpha_\text{bc}^\text{n},$ and $\alpha_\text{int}$ are the weights (penalties) corresponding to the loss terms associated with the \sloppy Dirichlet boundary condition, Neumann boundary condition, and interface conditions, respectively. $\text{MSE}_{\text{eq}},~\text{MSE}_{\text{bc}}^\text{n},~\text{MSE}_{\text{bc}}^\text{d},~\text{MSE}_{\text{int}}^\text{n},$ and $\text{MSE}_{\text{int}}^\text{d}$ are the mean-squared residual values evaluated for the governing equations, the Neumann boundary conditions, Dirichlet boundary conditions and the interface conditions, respectively, and are defined as
\begin{align}
   \begin{split}
    \text{MSE}_\text{eq} &= \displaystyle\sum_{\text{m}=1}^{2}\left( \frac{1}{\text{N}^\text{m}_{\Omega}}\sum_{\text{j}=1}^{\scriptscriptstyle{\text{N}^\text{m}_{\Omega}}}(\nabla\cdot({\kappa_\text{m}}_\text{j}\nabla {u^{\theta}_\text{m}}_\text{j}) + {f_\text{m}}_\text{j})^2\right), \\
    \text{MSE}^\text{d}_\text{bc} &= \displaystyle\sum_{\text{m}=1}^{2}\left(\frac{1}{\text{N}^\text{m}_{\partial\Omega^\text{d}}}\sum_{\text{j}=1}^{\scriptscriptstyle{\text{N}^\text{m}_{\partial\Omega^\text{d}}}} ({u^\theta_\text{m}}_\text{j}-{\Lambda^\text{d}_\text{m}}_\text{j})^2\right),\\
    \text{MSE}^\text{n}_\text{bc}&= \displaystyle\sum_{\text{m}=1}^{2}\left(\frac{1}{\text{N}^\text{m}_{\partial\Omega^\text{n}}}\sum_{\text{j}=1}^{\scriptscriptstyle{\text{N}^\text{m}_{\partial\Omega^\text{n}}}} ({{\kappa_\text{m}}_\text{j} \nabla {u_\text{m}^\theta}_\text{j}\cdot\mathbf{n_0}}-{\Lambda^\text{n}_\text{m}}_\text{j})^2\right),\\
    \text{MSE}^\text{d}_\text{ic} &= \frac{1}{\text{N}_{\Gamma_\text{int}}}\sum_{\text{j}=1}^{\scriptscriptstyle{\text{N}_{\Gamma_\text{int}}}}\ (\llbracket u^\theta_\text{j} \rrbracket_\text{int} - p_\text{j})^2,\\
    \text{MSE}^\text{n}_\text{ic} &= \frac{1}{\text{N}_{\Gamma_\text{int}}}\sum_{\text{j}=1}^{\scriptscriptstyle{\text{N}_{\Gamma_\text{int}}}}(\llbracket {\kappa_\text{i}\nabla u^{\theta}_\text{j}}\rrbracket_\text{int}\cdot{\mathbf{n_2}}-q_\text{j})^2,
  \end{split}
  \label{eq:MSE_equation_ipinns}
\end{align}
where m denotes the subdomain number and $\text{N}^\text{m}_{\Omega}, \text{N}^\text{m}_{\partial\Omega^\text{d}},$ and $\text{N}^\text{m}_{\partial\Omega^\text{n}}$ are the number of collocation points over which the residuals of the governing equations, Dirichlet and Neumann boundary conditions are respectively evaluated for each subdomain. The other symbols have the same definitions as before and are self-explanatory. For problems with multiple subdomains and more than one interface, extending the above equations is straightforward and is omitted here for conciseness. 

\subsection{Adaptive I-PINNs (AdaI-PINNs)}\label{sec:adaptive_ipinns}
In the I-PINNs framework of Sarma et al.~\cite{sarma2024ipinns}, different AFs are prescribed across each subdomain and are predefined at training. In AdaI-PINNs, we relax this constraint by integrating the concept of \textit{adaptive activation functions} (AAFs) described in Jagtap et al.~\cite{jagtap2020adaptive} into I-PINNs. This involves modifying the I-PINNs framework so that, at any layer s in the m-th subdomain, the function \(f_\text{s}\) is defined as:
\[
f_i(\mathbf{x}_\text{m,s}, \boldsymbol{\theta}_\text{s}, a_\text{m}) = \sigma \left(n a_\text{m} (\mathbf{w}_s^T \mathbf{x}_\text{m,s} + \mathbf{b}_\text{s})\right).
\]
Here, \(a_\text{m}\) is the AAF parameter for the AF \(\sigma(\odot)\) in the subdomain m, which is trainable. Thus, for a problem with M subdomains, the machine learns the AAF parameter set \(\mathbf{a} = [a_1, a_2, \ldots, a_\text{M}]\) along with the common set of weights and biases \(\boldsymbol{\theta}\) for all the networks. Technically speaking, this parameter defines the slope of the AF. Hence, the AFs for the neural networks across the sub-domains vary in their slopes, which are learned by the machine. Additionally, we introduce a new hyperparameter \(n\), a scaling factor that accelerates convergence towards the global minima. 

\begin{figure}[!hbt]
\begin{centering}
\includegraphics[width=0.8\textwidth]{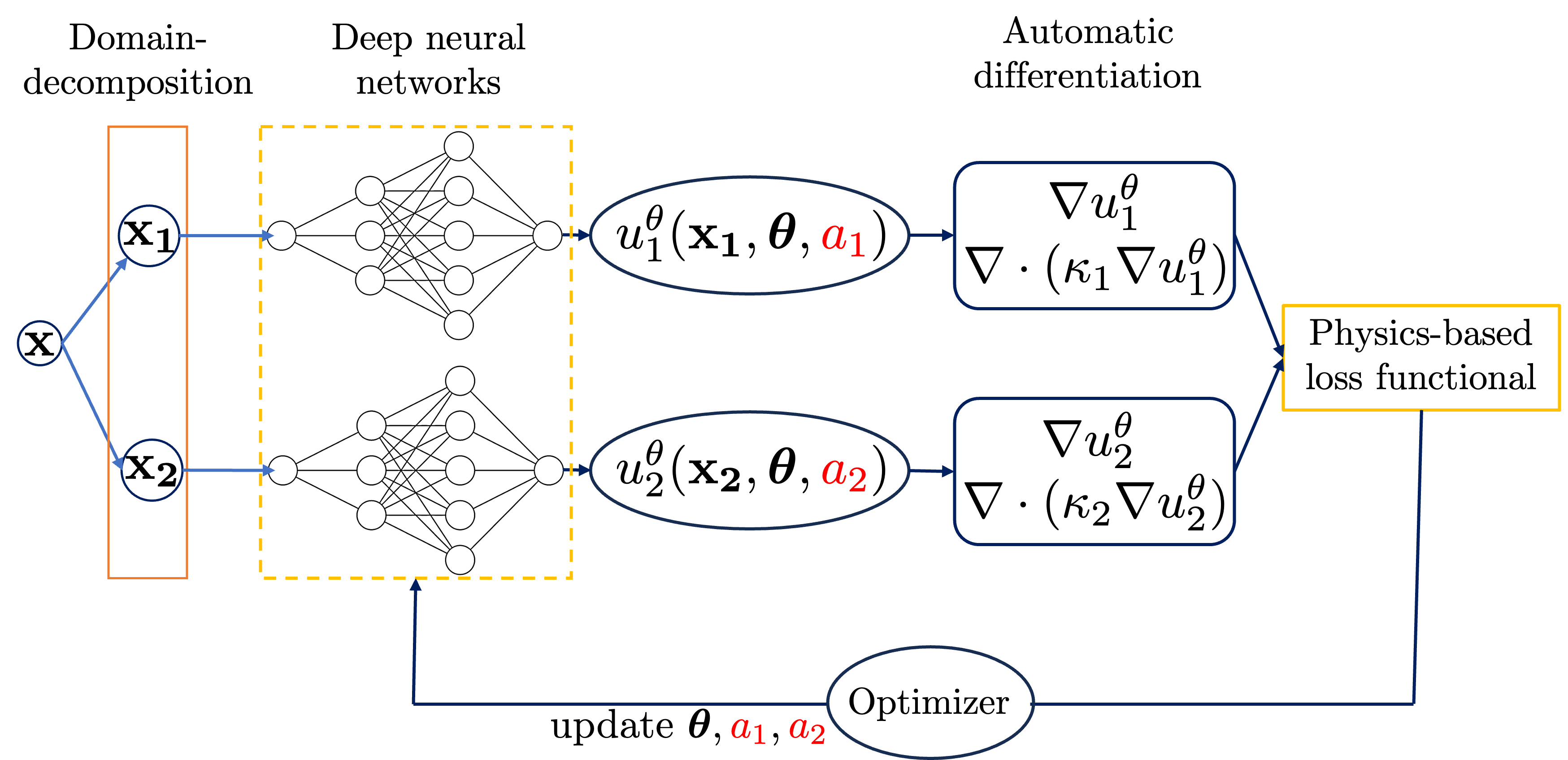}
\par\end{centering}
\caption{Schematic of the architecture of Adaptive Interface-PINNs (AdaI-PINNs) for a model problem consisting of a domain having two regions separated by an interface.}\label{fig:aipinns_schematic}
\end{figure}

It is important to note that the introduction of the AAF parameter \(\mathbf{a}\), as well as the scalable hyperparameter \(n\), does not change the structure of the previously defined loss functional. Therefore, the loss functional remains the same as defined in Eq. \eqref{eq:loss_function_ipinns}. Figure~\ref{fig:aipinns_schematic} is a schematic of the AdaI-PINNs framework for approximating the solution to the model problem, as depicted in Eqs.~\eqref{eq:Model_int_eqn}-\eqref{eq:Model_int_cond}.

\section{Numerical Experiments}\label{sec_experiments}

In this section, we evaluate the efficacy of the proposed AdaI-PINNs method by applying it to representative elliptic interface problems in 1D, 2D, and 3D. The AAFs employed for our experiments encompass a range of choices, including \textit{Sigmoid, Swish, Tanh, Softmax, GELU}, and \textit{Mish}. Some popular choices such as \textit{ReLU} and \textit{ELU} are not included in the experiments as they are known suffer from the vanishing gradient problem for second or higher order differential equations~\cite{jagtap2020adaptive, leng2022compatibility}. We evaluate the performance of the proposed AdaI-PINNs methodology through a comparison with I-PINNs (utilizing alternating activation functions). The root mean square error (RMSE) metric is used to compare the numerical accuracy of the experiments, while for quantifying the relative computational cost of an experiment, we define the metric $cost$ as:  $C = t_\text{m}/t_\text{AIP}$, where $t_\text{m}$ represents the total training time for the method under consideration, and $t_\text{AIP}$ represents the total training time for AdaI-PINNs. To gain insights into the convergence profiles, we analyze the variation of the total loss of each model with increasing iterations. The numerical experiments detailed in this study are conducted using functions from the Python JAX library \cite{jax2018github}. All optimizations are carried out using the Adam optimizer \cite{kingma2014adam} with an initial learning rate of $5\times10^{-3}$. Additionally, all parameters are initialized using the Xavier initialization scheme \cite{glorot2010understanding}, while the AAF parameters ($a_\text{m}$'s) were initialized to 0.5, and the scalable hyperparameter $n$ was fixed at 10 for all examples based on trial and error. It is worth highlighting that all the hyper-parameters are chosen through trial-and-error and are not necessarily optimal. Automated hyper-parameter tuning may further improve the accuracy of both the methods. 

\subsection{1D Example}\label{subsec_1d_examples}

We begin by considering the Poisson's equation in the domain $\Omega = [0,1]$ with four material interfaces specified at $x=0.2$, $x=0.4$, $x=0.6$, and $x=0.8$.
These interfaces partition the domain into five non-overlapping subdomains $\Omega_1 = [0,0.2],~\Omega_2 = [0.2,0.4]$, $\Omega_3 = [0.4,0.6]$, $\Omega_4 = [0.6,0.8]$, and $\Omega_5=[0.8,1]$, such that $\Omega=\Omega_{1}\cup\Omega_{2}\cup\Omega_{3}\cup\Omega_{4}\cup\Omega_{5}$. The governing PDE in each sub-domain $\Omega_\text{m}$ (for m = 1, 2, 3, 4, 5), the boundary and interface conditions are
\begin{align}\label{eq_1d_possions_four_int}
  \begin{split}
      \frac{d}{dx}\left({\kappa_\text{m}} \frac{d{u_\text{m}}}{dx}\right)&= - 1 \quad \text{in} ~ \Omega_\text{m}, \\
       u_{1} &= 0  \quad \text{at}~x=0, \\
       u_{5} &= 0  \quad \text{at}~x=1, \\
       \llbracket u \rrbracket &=0, ~\text{at}~ x={0.2, 0.4, 0.6, 0.8}, \\
       \left \llbracket {\kappa\frac{du}{dx}}  \right \rrbracket &=0 ~\text{at}~ x={0.2, 0.4, 0.6, 0.8}. 
  \end{split}
\end{align}
The material constants are discontinuous across each interface and are given as $\kappa_1=1$, $\kappa_2=0.25$, $\kappa_3=0.9$, $\kappa_4=0.1$, and $\kappa_5=0.8$. The closed-form analytical solution to the set of Eqs.~\eqref{eq_1d_possions_four_int} in each subdomain $\Omega_\text{m}$ is a quadratic function of x, such that 
\begin{equation}
    u_\text{m}=c^0_\text{m}x^2 + c^1_\text{m}x + c^2_\text{m},
\end{equation}
where the coefficients $c^0_\text{m}, c^1_\text{m}$ and $c^2_\text{m}$ are functions of the material constants $\kappa_1, \kappa_2, \kappa_3$, $\kappa_4$ and $\kappa_5$. The explicit definitions of these coefficients are omitted here for conciseness.

\begin{figure}[!hbt]
    \centering
    \begin{subfigure}[b]{0.48\textwidth}
        \centering
        \includegraphics[width=\textwidth]{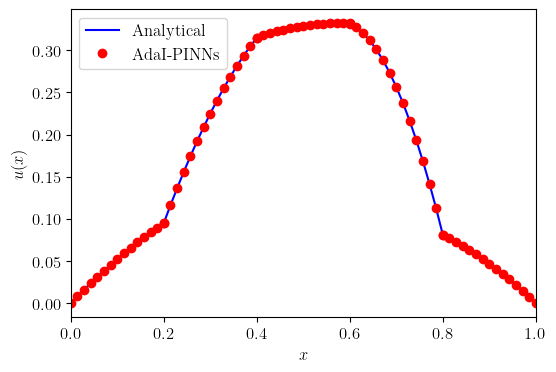}
        \caption{}
        \label{fig_1d_adap_result}
    \end{subfigure}
    \begin{subfigure}[b]{0.49\textwidth}
        \centering
        \includegraphics[width=\textwidth]{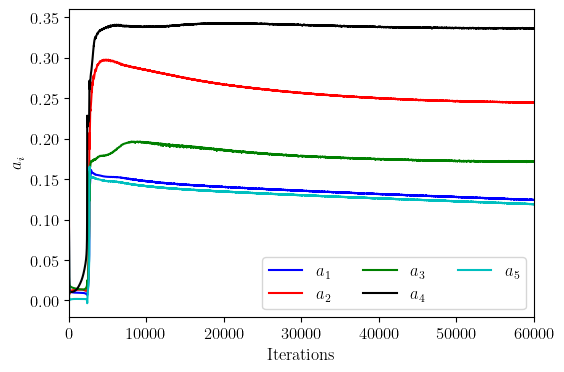}
        \caption{}
        \label{fig_1d_adap_trained_as}
    \end{subfigure}    
    \caption{(a) Approximation to Eq.~\eqref{eq_1d_possions_four_int} by the AdaI-PINNs method compared to the closed-form analytical solution, (b) the iterative variation of the parameters $a_\text{m}$ corresponding to each sub-domain $\Omega_\text{m}$ within the adaptive tanh activation function.}
    \label{fig_1d_adap_result_trained_as}
\end{figure}

Figure~\ref{fig_1d_adap_result} compares the AdaI-PINNs approximation (red dots) to Eq.\eqref{eq_1d_possions_four_int} with the closed-form analytical solution (blue solid line). Figure~\ref{fig_1d_adap_trained_as} shows the variation of the AAF parameters with the number of iterations. The hyper-parameters utilized for training the adaptive I-PINNs model are outlined in the first column of Table~\ref{table_hyperparameter_1d_four_int}. It is observed that the AdaI-PINNs approximation shows excellent agreement with the analytical solution throughout the domain, as well as at the interfaces, resulting in an RMSE of $1.05\times 10^{-4}$. By contrast, I-PINNs achieve an RMSE of only $\mathcal{O}(10^{-2})$ on using identical hyper-parameters (as detailed in column 2 of Table~\ref{table_hyperparameter_1d_four_int}). To achieve the same accuracy as AdaI-PINNs, I-PINNs require approximately three times as many iterations (200,000 versus 60,000), and consequently they are approximately twice (1.94 versus 1) as expensive as AdaI-PINNs (see column 3 of Table~\ref{table_hyperparameter_1d_four_int}).  This observation is supported by Figure~\ref{fig_1d_adap_trained_as}, where AdaI-PINNs (with training stopped at 60,000 iterations) exhibit significantly faster and more effective convergence compared to I-PINNs (trained to 200,000 iterations). The latter tends to get trapped in local minima and only breaks free after approximately 50,000 iterations, whereas AdaI-PINNs exhibit more monotonic convergence behavior. 

Table~\ref{table_as_rmse_cost_1d_four_int} provides an overview of the RMSE and computational costs associated with training the AdaI-PINNs framework using various AFs. Notably, the accuracy of AdaI-PINNs appears insensitive to activation functions, consistently yielding RMSEs of $\mathcal{O}(10^{-4})$. However, when comparing between AFs, it is clear that adaptive tanh and adaptive sigmoid functions stand out as the most computationally efficient options. These results were obtained using the hyperparameters outlined in column 1 of Table~\ref{table_hyperparameter_1d_four_int}, with variations limited to the AFs. Additionally, Table~\ref{table_as_rmse_cost_1d_four_int} also includes the trained AAF parameters $a_\text{m}$ for each case. This is to be contrasted with the results reported in Sarma et al.~\cite{sarma2024ipinns}, where the results from I-PINNs exhibited more sensitivity to AFs. Moreover, while a more complex layer architecture could be used in I-PINNs to obtain the same accuracy as AdaI-PINNs, it led to a higher computational cost underscoring the benefits of the AdaI-PINNs approach in achieving both accuracy and computational efficiency simultaneously.

\begin{table}[!hbt]
  \begin{center}
    \caption{Hyper-parameters used in training the proposed AdaI-PINNs model, as well as I-PINNs model for approximating the solution to 1D problem (Eq.~\eqref{eq_1d_possions_four_int}).}
    \label{table_hyperparameter_1d_four_int}
    \small
    \begin{tabular}{c|c|c|c}
      \toprule 
      \textbf{Hyper-parameters} & \textbf{AdaI-PINNs} &  \textbf{I-PINNs} & \textbf{I-PINNs} \\
      & & &\textbf{(More iterations)}\\
      \midrule 
      \# of hidden layers & 3 & 3 & 3\\
      \# of neurons per hidden layer & 10 & 10 & 10\\
      \# of training points ($N_\Omega$) & 131 & 131 & 131 \\
      $\alpha_\text{int}$ at each interface & 5 & 5 & 5 \\
      $\alpha_\text{bc}$ at each boundary & 10 & 10 & 10\\
      Activation functions & adaptive tanh & tanh--swish-- & tanh--swish--\\
      & & tanh--swish--tanh & tanh--swish--tanh \\
      Iterations & 60000 & 60000 & 200000 \\
      RMSE & $1.05\times10^{-4}$ & $5.07\times10^{-2}$ & $2.42\times10^{-4}$ \\
      Cost  & 1 & 0.78 & 1.94\\
      \bottomrule
    \end{tabular}
  \end{center}
\end{table}

\begin{figure}[!hbt]\label{fig_1d_four_int_ipinns_adapipinns_convergence}
    \centering
    \includegraphics[width=0.55\textwidth]{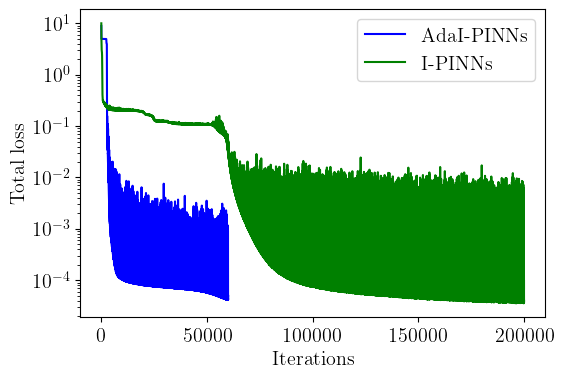}
    \caption{The iterative variation of total loss by both AdaI-PINNs and I-PINNs, for the 1D problem outlined in Eq.~\eqref{eq_1d_possions_four_int}. The training of the AdaI-PINNs model was clipped at 60,000 iterations as it reached convergence very quickly, while I-PINNs was left to train for 200000 iterations.}
    \label{fig:enter-label}
\end{figure}

\begin{table}[!hbt]
  \begin{center}
    \caption{RMSE, computational cost, and the trained parameters $a_\text{m}$'s obtained by the AdaI-PINNs method for approximating the solution to the 1D problem, given by Eq.~\eqref{eq_1d_possions_four_int} across different adaptive activation functions. The framework was trained for 60,000 iterations for each experiment.}    
    \label{table_as_rmse_cost_1d_four_int}
    \small
    \begin{tabular}{c|c|c|c|c|c|c|c}
      \toprule 
      \textbf{AAF} & \textbf{\textit{a}}$\mathbf{_1}$ &  \textbf{\textit{a}}$\mathbf{_2}$ & \textbf{\textit{a}}$\mathbf{_3}$ & \textbf{\textit{a}}$\mathbf{_4}$ & \textbf{\textit{a}}$\mathbf{_5}$ & \textbf{RMSE} & \textbf{Cost}\\
      \midrule 
      Tanh & $0.12423$ & $0.24443$ & $0.17177$ & $0.33630$ & $0.11897$ &$1.05 \times 10^{-4}$ & 1\\
      Swish & $0.15743$ & $0.24723$ & $0.19361$ & $0.28725$ & $0.15052$ & $2.97 \times 10^{-4}$ & 1.41 \\
      Sigmoid & $0.27868$ & $0.64592$ & $0.38892$ & $0.93277$ & $0.26395$ & $2.15 \times 10^{-4}$ & 1.03\\
      Softmax & $-0.29144$ & $-0.51919$ & $-0.34869$ & $-0.54655$ & $-0.27954$ & $7.05 \times 10^{-4}$ & 1.68\\
      GELU & $0.14182$ & $0.24466$ & $0.17210$ & $0.27477$ & $0.13580$ & $6.37 \times 10^{-4}$ & 1.49\\
      Mish & $0.15148$ & $0.23800$ & $0.18388$ & $0.30368$ & $0.14516$ & $2.76 \times 10^{-4}$ & 1.42\\
      \bottomrule
    \end{tabular}
  \end{center}
\end{table}

\subsection{Two-dimensional (2D) Example}\label{sec:2d_examples}
We now consider Poisson's equation defined in a 2D rectangular domain \(\Omega = [0,1.7] \times [0,1]\), with four inclusions describing the letters `I' (\(\Omega_2\)), `I' (\(\Omega_3\)), `T' (\(\Omega_4\)), `M' (\(\Omega_5\)), and the matrix in the background (\(\Omega_1\)), such that \(\Omega = \Omega_{1} \cup \Omega_{2} \cup \Omega_{3} \cup \Omega_{4} \cup \Omega_{5}\). The governing PDE in each subdomain $\Omega_\text{m}$ (for m = 1, 2, 3, 4, 5) is given by 
\begin{align}
  \begin{split}
    \nabla\cdot(\kappa_\text{m}\nabla u_\text{m}) &= 1 \hspace{1.11cm}\qquad \text{in}~~\Omega_\text{m},\\
    u_\text{m} &= \Lambda^{\text{d}}_\text{m} \hspace{0.7 cm}\qquad \text{on}~~\partial\Omega^\text{d}_\text{m},\\
    \llbracket u \rrbracket  &= p \qquad \hspace{1.05 cm}\text{on}~\Gamma_\text{int},\\
    \llbracket \kappa \nabla{u}\rrbracket\cdot{\mathbf{n}_\textbf{2}} &= q \qquad \hspace{1.07 cm} \text{on}~\Gamma_\text{int},
  \end{split}
  \label{eq:2d_iitm_equation}
\end{align}
where $\partial\Omega^\text{d}_\text{1} = \left\{\mathbf{x}: x = 1.7\cup y = 0 \right\}$ and $\partial\Omega^\text{d}_\text{2} = \left\{\mathbf{x}: x= 0 \cup  y = 1 \right\}$ are the parts of the external boundary with prescribed Dirichlet boundary conditions. Considering $\kappa_1 = 1/4$, $\kappa_2 = 1/6$, $\kappa_3 = 1/10$, $\kappa_4 = 1/14$ and $\kappa_5 = 1/3$, the set of Eqs.~\eqref{eq:2d_iitm_equation} have a closed-form analytical solution given as 
\begin{align}
\begin{split}
    u(\textbf{x})&=
    \begin{cases}
      x^{2}+y^{2} &\mbox{in $\Omega_1$}\\
      3x^{2}+2y &\mbox{in $\Omega_2$}\\
      4x^{2}+y^{2} &\mbox{in $\Omega_3$}\\
      x^{2}+5y^{2} &\mbox{in $\Omega_4$}\\
      0.5x^{2}+y^{2} &\mbox{in $\Omega_5$}
    \end{cases}
\end{split}\label{eq:2d_analytical}
\end{align}
The quantities, $p$, $q$ and $\Lambda^\text{d}_\text{m}$ were extracted from Eq~\eqref{eq:2d_analytical}. 

\begin{figure}[!hbt]
    \centering
    \begin{subfigure}[b]{0.45\textwidth}
    \includegraphics[width=\textwidth]{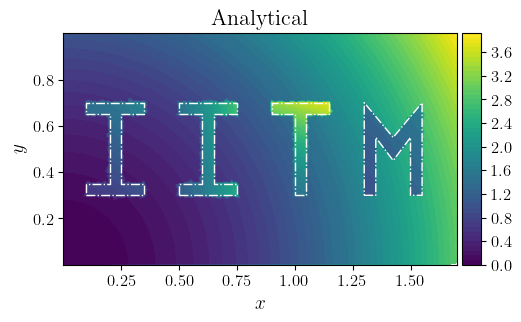}
    \caption{}
    \label{fig: 2d_analytical}
    \end{subfigure}\\
    \centering
    \begin{subfigure}[b]{0.45\textwidth}
        \centering
        \includegraphics[width=\textwidth]{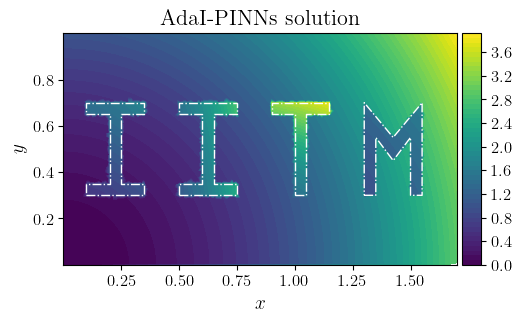}
        \caption{}
        \label{fig:2d_aipinns}
    \end{subfigure}
    \begin{subfigure}[b]{0.45\textwidth} 
        \centering
        \includegraphics[width=\textwidth, height = 4.6cm]{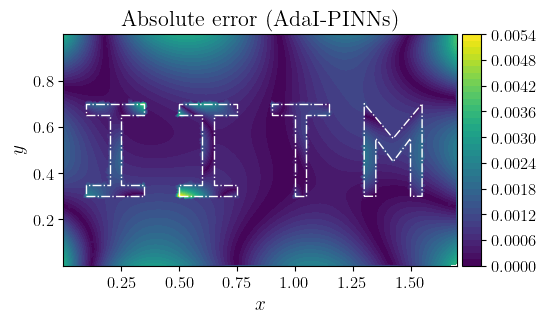}
        \caption{}
        \label{fig:2d_aipinns_error} 
    \end{subfigure}\\
    \caption{(a) Contour plot of the analytical solution to Eq~\eqref{eq:2d_iitm_equation}, (b) approximation to the solution by AdaI-PINNs, and, (c) absolute error plot of the approximation obtained AdaI-PINNs (RMSE = $3.48 \times 10^{-6}$). The interfaces are demarcated by the white dash-dot lines.}
    \label{fig:2d_aipinns_full}
\end{figure}

\begin{figure}[!hbt]
    \centering
    \begin{subfigure}[b]{0.48\textwidth}
        \centering
        \includegraphics[width=\textwidth]{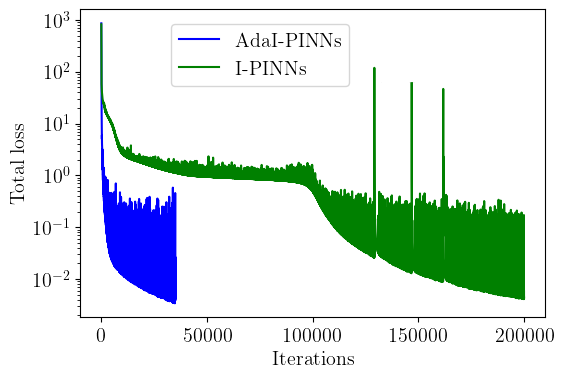}
        \caption{}
        \label{fig_2d_convergence}
    \end{subfigure}
    \begin{subfigure}[b]{0.47\textwidth}
        \centering
        \includegraphics[width=\textwidth]{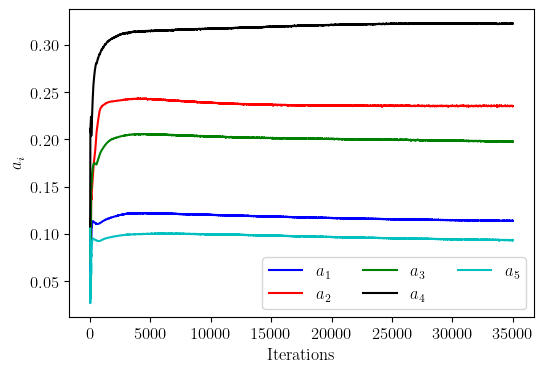}
        \caption{}
        \label{fig_2d_ais_convergence}
    \end{subfigure}    
    \caption{The iterative variation of (a) total loss by both AdaI-PINNs and I-PINNs, and (b) AAF parameter $a_\text{i}$ for each subdomain $\Omega_\text{i}$ for the AdaI-PINNs framework, for the 2D inclusion problem given by Eq.~\eqref{eq:2d_iitm_equation}. The training of the AdaI-PINNs model was clipped at 35,000 iterations as it reached convergence very quickly, while I-PINNs was let to train for 200000 iterations.}
    \label{fig:2d_iitm_convergence_full}
\end{figure}

A contour plot depicting the solution to Eqs.~\eqref{eq:2d_iitm_equation}, along with the corresponding absolute error plot, approximated by the AdaI-PINNs framework, is presented in Figure~\ref{fig:2d_aipinns_full}. The hyperparameters used to achieve this solution are listed in the first column of Table~\ref{table_hyperparameter_2d_iitm}. The model was trained for 60,000 iterations, resulting in an RMSE of $\mathcal{O}(10^{-6})$. Using the same hyperparameters, the I-PINNs model was also run, utilizing the swish activation function in the inclusions and tanh in the solid matrix. After 60,000 iterations, the I-PINNs model achieved an RMSE of only $\mathcal{O}(10^{-3})$. To reach the same level of accuracy as AdaI-PINNs, i.e., $\mathcal{O}(10^{-6})$, the I-PINNs model required 200,000 iterations (refer to the third column of Table~\ref{table_hyperparameter_2d_iitm}), which is approximately twice as expensive computationally as AdaI-PINNs. This is further demonstrated in Figure~\ref{fig_2d_convergence}, which clearly shows that AdaI-PINNs converge significantly faster than I-PINNs, achieving such an RMSE with much fewer iterations. Additionally, the AAF parameters converge to their steady-state values relatively quickly, as illustrated in Figure~\ref{fig_2d_ais_convergence}.

Table~\ref{table_as_rmse_cost_2d} provides an overview of the RMSE and computational cost associated with training the AdaI-PINNs framework to approximate the solution to Eq.~\eqref{eq:2d_iitm_equation} using various activation functions (AFs). Once again, we notice that the accuracy of AdaI-PINNs is relatively insensitive to the choice of AFs, with most yielding RMSEs of $\mathcal{O}(10^{-5})$ and $\mathcal{O}(10^{-6})$. Interestingly, as in the 1D case, sigmoid and tanh were the most computationally efficient AFs. These results were obtained using the hyperparameters outlined in column 1 of Table~\ref{table_hyperparameter_2d_iitm}, with variations limited to the AFs. Table~\ref{table_as_rmse_cost_2d} also lists the magnitudes of the trained AAF parameters $a_\text{i}$ for each case.

\begin{table}[!hbt]
  \begin{center}
    \caption{Hyper-parameters used in training the proposed AdaI-PINNs model, as well as I-PINNs model for approximating the solution to 2D inclusion problem (Eq.~\eqref{eq:2d_iitm_equation}).}
    \label{table_hyperparameter_2d_iitm}
    \small
    \begin{tabular}{c|c|c|c}
      \toprule 
      \textbf{Hyper-parameters} & \textbf{AdaI-PINNs} &  \textbf{I-PINNs} & \textbf{I-PINNs} \\
      & & &\textbf{(More iterations)}\\
      \midrule 
      \# of hidden layers & 3 & 3 & 3\\
      \# of neurons per hidden layer & 20 & 20 & 20\\
      \# of training points ($N_\Omega$) & 3679 & 3679 & 3679 \\
      $\alpha_\text{int}$ at each interface & 25 & 25 & 25 \\
      $\alpha_\text{bc}$ at each boundary & 20 & 20 & 20\\
      Activation functions & adaptive tanh & tanh--swish & tanh--swish\\
      Iterations & 60,000 & 60,000 & 200,000 \\
      RMSE & $3.48\times10^{-6}$ & $1.93\times10^{-3}$ & $5.56\times10^{-6}$ \\
      Cost  & 1 & 0.49 & 2.06\\
      \bottomrule
    \end{tabular}
  \end{center}
\end{table}

\begin{table}[!hbt]
  \begin{center}
    \caption{RMSE, computational cost, and the trained parameters $a_\text{m}$'s obtained by the AdaI-PINNs method for approximating the solution to the 2D problem, given by Eq.~\eqref{eq:2d_iitm_equation} across different adaptive activation functions. The framework was trained for 60,000 iterations for each expriment.}    
    \label{table_as_rmse_cost_2d}
    \small
    \begin{tabular}{c|c|c|c|c|c|c|c}
      \toprule 
      \textbf{AAF} & \textbf{\textit{a}}$\mathbf{_1}$ &  \textbf{\textit{a}}$\mathbf{_2}$ & \textbf{\textit{a}}$\mathbf{_3}$ & \textbf{\textit{a}}$\mathbf{_4}$ & \textbf{\textit{a}}$\mathbf{_5}$ & \textbf{RMSE} & \textbf{Cost}\\
      \midrule 
      Tanh & $0.11379$ & $0.23504$ & $0.19759$ & $0.32251$ & $0.09323$ &$3.48 \times 10^{-6}$ & 1\\
      Swish & $0.13489$ & $0.28312$ & $0.25126$ & $0.19883$ & $0.11978$ & $3.24 \times 10^{-6}$ & 1.55 \\
      Sigmoid & $0.20294$ & $0.79904$ & $0.60419$ & $0.42119$ & $-0.31878$ & $1.59 \times 10^{-5}$ & 0.96\\
      Softmax & $-0.31080$ & $0.40909$ & $0.55511$ & $0.87228$ & $-0.52780$ & $1.37 \times 10^{-6}$ & 1.78\\
      GELU & $0.11873$ & $0.25544$ & $0.22200$ & $0.17704$ & $0.10606$ & $1.50 \times 10^{-6}$ & 1.58\\
      Mish & $0.12410$ & $0.26738$ & $0.23596$ & $0.18745$ & $0.11126$ & $2.25 \times 10^{-5}$ & 1.48\\
      \bottomrule
    \end{tabular}
  \end{center}
\end{table}

\subsection{Three-dimensional (3D) Example}\label{sec:3d_examples}
As a last example, we compare the performance of AdaI-PINNs with I-PINNs by solving Poisson’s equation defined in the 3D cubic computational domain $\Omega = [-1,1]\times [-1,1]\times [-1,1]$, with eight spherical subdomains of radius $r = 0.3$ defined as
\clearpage
\begin{align*}
    \Omega_2 &= \{\mathbf{x} : (x + 0.5)^2 + (y + 0.5)^2 + (z + 0.5)^2 \le 0.3^2\},\\
    \Omega_3 &= \{\mathbf{x} : (x + 0.5)^2 + (y - 0.5)^2 + (z + 0.5)^2 \le 0.3^2\},\\
    \Omega_4 &= \{\mathbf{x} : (x - 0.5)^2 + (y + 0.5)^2 + (z + 0.5)^2 \le 0.3^2\},\\
    \Omega_5 &= \{\mathbf{x} : (x - 0.5)^2 + (y - 0.5)^2 + (z + 0.5)^2 \le 0.3^2\},\\
    \Omega_6 &= \{\mathbf{x} : (x + 0.5)^2 + (y + 0.5)^2 + (z - 0.5)^2 \le 0.3^2\},\\
    \Omega_7 &= \{\mathbf{x} : (x + 0.5)^2 + (y - 0.5)^2 + (z - 0.5)^2 \le 0.3^2\},\\
    \Omega_8 &= \{\mathbf{x} : (x - 0.5)^2 + (y + 0.5)^2 + (z - 0.5)^2 \le 0.3^2\},\\
    \Omega_9 &= \{\mathbf{x} : (x - 0.5)^2 + (y - 0.5)^2 + (z - 0.5)^2 \le 0.3^2\}.
\end{align*}

The rest of the domain is defined as, $\Omega_1 = \Omega \setminus \overline{(\Omega_2 \cup \Omega_3 \cup \Omega_4 \cup \Omega_5 \cup \Omega_6 \cup \Omega_7 \cup \Omega_8 \cup \Omega_9)}$. The equations of the interfaces are
\begin{equation}
\begin{aligned}
    \Gamma_\text{int} = \{ \textbf{x} : & \ \psi(\textbf{x}) = 0, \\
    & \ \psi(\textbf{x}) = \min_{0 \leq k \leq 7} \left( \sqrt{(x-x_k)^2+(y-y_k)^2+(z-z_k)^2} \right) - 0.3 \}, \\
    & \ \text{where} \ (x_k, y_k, z_k) = ((-1)^{\lfloor{k/4}\rfloor} \times 0.5, 
    (-1)^{\lfloor{k/2}\rfloor} \times 0.5, 
    (-1)^{\lfloor{k}\rfloor} \times 0.5), \ 0 \leq k \leq 7.
\end{aligned}
\end{equation}
where $\lfloor\odot\rfloor$ represents the floor function for the quantity $\odot$. The model problem in these nine regions is defined as
\begin{align}
  \begin{split}
    \nabla\cdot(\kappa_\text{m}\nabla u_\text{m}) &= 1 \hspace{1 cm}\qquad \text{in}~~\Omega_\text{m},\\
    u_\text{5} &= \Lambda^\text{d} \hspace{0.7 cm}\qquad \text{on}~~\partial\Omega^\text{d}_\text{1},\\
    \llbracket u \rrbracket_\text{j}  &= \lambda_\text{j} \qquad \hspace{0.8 cm}\text{on}~\Gamma_\text{int}^\text{j},\\
    \llbracket \kappa \nabla{u}\rrbracket_\text{j}\cdot{\mathbf{n}_\textbf{j}} &= \chi_\text{j} \qquad \hspace{.75 cm} \text{on}~\Gamma_\text{int}^\text{j},
  \end{split}
  \label{eq:3d_ball_equation}
\end{align}
where $\partial\Omega^\text{d}_\text{1} = \left\{\mathbf{x}: x = 0 \cup x = 1 \cup y = 0 \cup y = 1 \cup z = -1 \cup z = 1 \right\}$ is the external boundary with Dirichlet boundary conditions. Considering the material constants, $\kappa_1 = \frac{1}{6}$, $\kappa_2 = \frac{1}{8}$, $\kappa_3 = \frac{1}{14}$, $\kappa_4 = \frac{1}{16}$, $\kappa_5 = \frac{1}{10}$, $\kappa_6 = \frac{1}{2}$, $\kappa_7 = \frac{1}{12}$, $\kappa_8 = \frac{1}{9}$, and, $\kappa_9 = \frac{1}{18}$, the set of Equations \eqref{eq:3d_ball_equation} have a closed-form analytical solution given as
\begin{align}
\begin{split}
    u(\textbf{x})&=
    \begin{cases}
      x^{2}+y^{2}+z^2 &\mbox{in $\Omega_1$}\\
      3x^{2}+2y+z^2 &\mbox{in $\Omega_2$}\\
      4x^{2}+y^{2}+2z^2 &\mbox{in $\Omega_3$}\\
      x^{2}+5y^{2}+2z^2 &\mbox{in $\Omega_4$}\\
      x^{2}+3y^2+z^2 &\mbox{in $\Omega_5$}\\
      2x+y^2+2z &\mbox{in $\Omega_6$}\\
      5x^{2}+2y+z^2 &\mbox{in $\Omega_7$}\\
      2x^{2}+2y^2+0.5z^2 &\mbox{in $\Omega_8$}\\
      3x^{2}+5y^2+z^2 &\mbox{in $\Omega_9$}
    \end{cases}
\end{split}\label{eq:3d_balls_anal}
\end{align}
The known boundary data $\Lambda^{\text{d}} $ at the Dirichlet boundaries and the jumps in primary variable and flux at the interfaces ($\Gamma_\text{int}^j$, j = 1 to 8), $\lambda_\text{j}$ and $\chi_\text{j}$ respectively, are prescribed according to Equation \eqref{eq:3d_balls_anal}.

\begin{figure}[!hbt]
\begin{centering}
\includegraphics[width=0.8\textwidth]{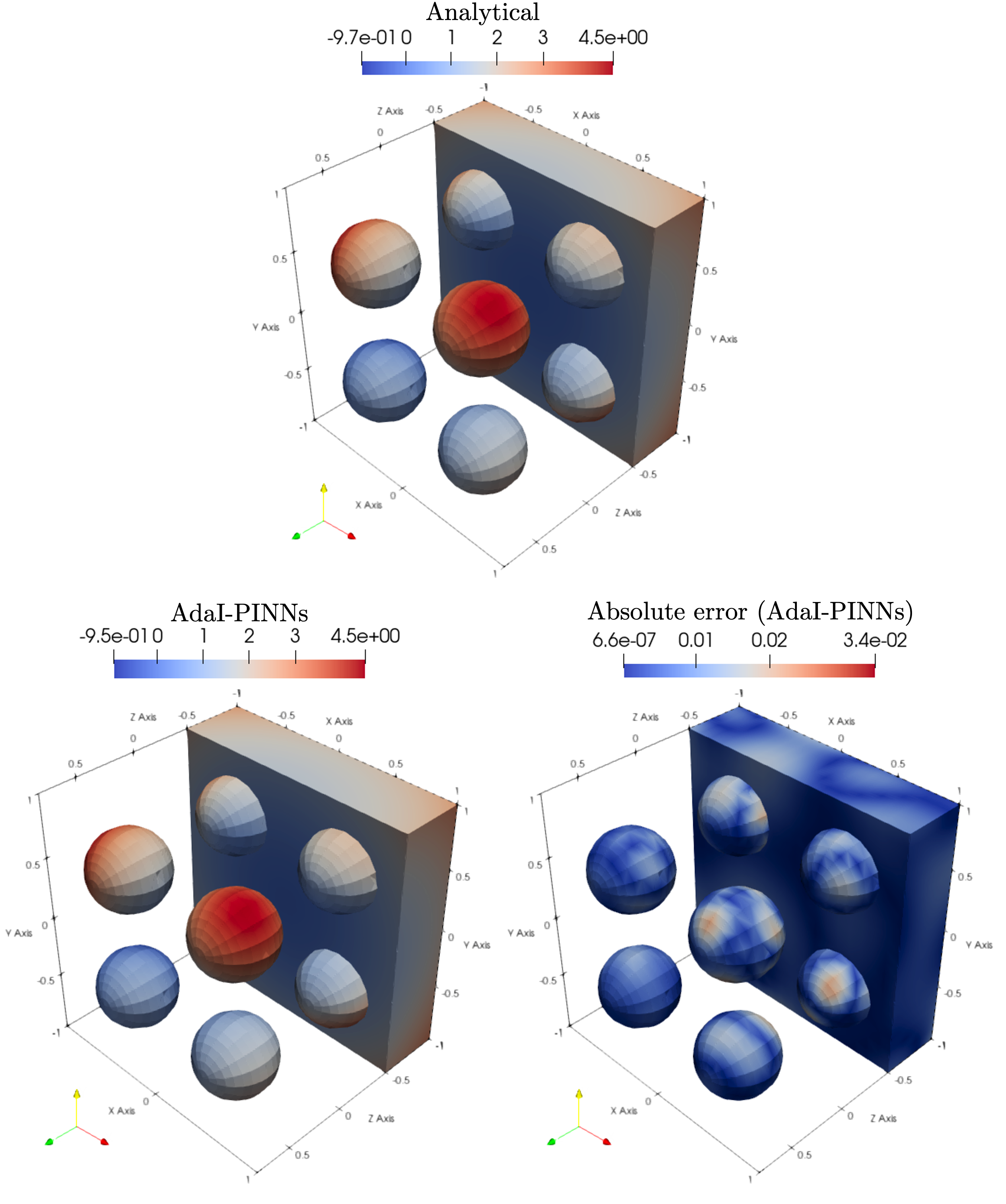}
\par\end{centering}
\caption{Contour plots of approximation to the solution to Equation~\eqref{eq:3d_ball_equation} by the AdaI-PINNs frameworks, and its corresponding absolute error plot. The domain is clipped at $z=-0.5$ to visualize the solution field in the interior of the cubical domain and along the eight interfaces.}\label{fig:3d_aipinns_full}
\end{figure}

A contour plot depicting the solution to Eq.~\ref{eq:3d_ball_equation}, along with the corresponding absolute error plot, is shown in Figure~\ref{fig:3d_aipinns_full}. The solution is obtained using the hyperparameters listed in the first column of Table~\ref{table_hyperparameter_3d_balls}. For enhanced clarity, the domain is clipped across the plane $z=-0.5$ to observe the variation of the solution field within the interior of the domain and along the interfaces. Using the same set of hyperparameters, the I-PINNs framework is also trained. The key difference is that AdaI-PINNs employ nine different adaptive sigmoid activation functions for the subdomains, whereas I-PINNs use sigmoid functions within the inclusions and the tanh function in the background matrix. Both frameworks were tested for 20,000 iterations. From Table~\ref{table_hyperparameter_3d_balls}, we observe that although I-PINNs are computationally cheaper to reach 20,000 iterations compared to AdaI-PINNs, I-PINNs only achieve an accuracy of $\mathcal{O}(10^{-2})$, while AdaI-PINNs achieve an accuracy of $\mathcal{O}(10^{-3})$. Furthermore, even if the iterations were increased to 250,000, I-PINNs still could not match the accuracy of AdaI-PINNs, yielding an RMSE of $\mathcal{O}(10^{-2})$ and being almost 6 times more computationally expensive (see column 3 of Table~\ref{table_hyperparameter_3d_balls}).

\begin{table}[!hbt]
  \begin{center}
    \caption{Hyper-parameters used in training the proposed adaptive I-PINNs model, as well as I-PINNs model for approximating the solution to 3D problem (Eq.~\eqref{eq:3d_ball_equation}).}
    \label{table_hyperparameter_3d_balls}
    \small
    \begin{tabular}{c|c|c|c}
      \toprule 
      \textbf{Hyper-parameters} & \textbf{AdaI-PINNs} &  \textbf{I-PINNs} & \textbf{I-PINNs} \\
      & & &\textbf{(More iterations)}\\
      \midrule 
      \# of hidden layers & 2 & 2 & 2\\
      \# of neurons per hidden layer & 50 & 50 & 50\\
      \# of training points ($N_\Omega$) & 3336 & 3336 & 3336 \\
      $\alpha_\text{int}$ at each interface & 50 & 50 & 50 \\
      $\alpha_\text{bc}$ at each boundary & 40 & 40 & 40\\
      Activation functions & adaptive sigmoid & swish--sigmoid & swish--sigmoid\\
      Iterations & 20000 & 20000 & 250000 \\
      RMSE & $5.47\times10^{-3}$ & $2.06\times10^{-2}$ & $1.37\times10^{-2}$ \\
      Cost  & 1 & 0.65 & 5.93\\
      \bottomrule
    \end{tabular}
  \end{center}
\end{table}

\begin{figure}[!hbt]
    \centering
    \begin{subfigure}[b]{0.48\textwidth}
        \centering
        \includegraphics[width=\textwidth]{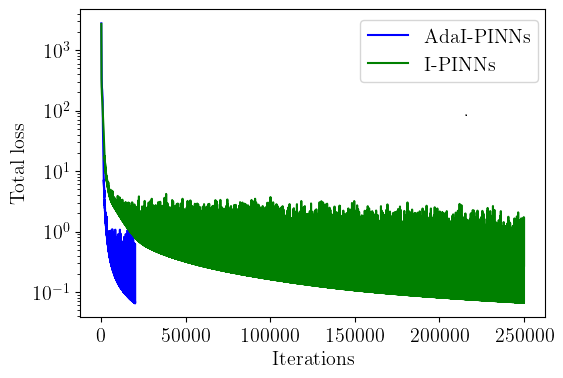}
        \caption{}
        \label{fig:3d_loss_convergence}
    \end{subfigure}
    \begin{subfigure}[b]{0.47\textwidth}
        \centering
        \includegraphics[width=\textwidth]{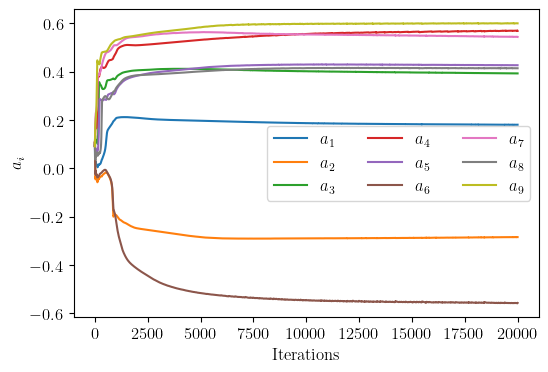}
        \caption{}
        \label{fig:3d_adap_convergence}
    \end{subfigure}    
    \caption{The iterative variation of (a) total loss by both AdaI-PINNs and I-PINNs, and (b) AAF parameter $a_\text{i}$ for each subdomain $\Omega_\text{i}$ for the AdaI-PINNs framework, for the 3D inclusion problem given by Eq.~\eqref{eq:3d_ball_equation}. The training of the AdaI-PINNs model was clipped at 20,000 iterations as it reached convergence very quickly, while I-PINNs was left to train for 250000 iterations.}
    \label{fig:3d_iitm_convergence_full}
\end{figure}

Figure~\ref{fig:3d_loss_convergence} shows the iterative variation of the total loss functions for both the AdaI-PINNs and I-PINNs models. Similar to the 1D and 2D examples, AdaI-PINNs converge much faster than I-PINNs, reaching loss values of magnitude $\mathcal{O}(10^{-2})$ in only 20,000 iterations (hence the training for AdaI-PINNs is stopped at 20,000 iterations), while I-PINNs take almost 250,000 iterations to reach a similar loss. This observation is further supported by Figure~\ref{fig:3d_adap_convergence}, which plots the iterative variation of the AAF parameter (sigmoid) $a_\text{m}$ for each sub-domain $\Omega_\text{m}$ during AdaI-PINNs training. It is observed that, similar to the 1D and 2D problems, the model quickly learns the optimal combination of $a_\text{m}$ values. Consequently, $a_\text{m}$s converge to stable values very quickly, at almost around 6,000 iterations. 

Similar to the previous numerical examples, we present the results of using various AAFs for the AdaI-PINNs framework. We attempt to approximate the solution to Eq.~\ref{eq:3d_ball_equation} and compare each activation function's RMSE and computational costs. The approximations in 3D are more sensitive to activation functions than those in 1D and 2D, as shown in Table~\ref{table:as_rmse_cost_3d}, though the accuracies remain comparable across all activation functions. All results were generated by training the AdaI-PINNs framework for 75,000 iterations using the hyperparameters listed in the first column of Table~\ref{table_hyperparameter_3d_balls}, with only the AAF parameters being different. All AAF choices result in RMSEs on the order of $\mathcal{O}(10^{-2})$ and $\mathcal{O}(10^{-3})$, with the sigmoid function being the most efficient in terms of both accuracy and computational cost.

\begin{table}[!hbt]
  \begin{center}
    \caption{RMSE, computational cost, and the trained parameters $a_\text{m}$'s obtained by the AdaI-PINNs method for approximating the solution to the 3D problem, given by Eq.~\eqref{eq:3d_ball_equation} across different adaptive activation functions. The framework was trained for 75,000 iterations for each experiment.}    
    \label{table:as_rmse_cost_3d}
    \scriptsize 
    \begin{tabular}{c|c|c|c|c|c|c|c|c|c|c|c}
      \toprule 
      \textbf{AAF} & \textbf{\textit{a}}$\mathbf{_1}$ &  \textbf{\textit{a}}$\mathbf{_2}$ & \textbf{\textit{a}}$\mathbf{_3}$ & \textbf{\textit{a}}$\mathbf{_4}$ & \textbf{\textit{a}}$\mathbf{_5}$ &
      \textbf{\textit{a}}$\mathbf{_6}$ &
      \textbf{\textit{a}}$\mathbf{_7}$ &
      \textbf{\textit{a}}$\mathbf{_8}$ &
      \textbf{\textit{a}}$\mathbf{_9}$ &\textbf{RMSE} & \textbf{Cost}\\
      \midrule 
      Sigmoid & $0.1696$ & $-0.2887$ & $0.3823$ & $0.5651$ & $0.4069$ & $-0.5628$ & $0.5197$ & $0.4095$ & $0.5857$ &$4.32\times10^{-3}$ & 1\\
      Tanh & $0.1809$ & $0.1426$ & $0.2742$ & $0.3402$ & $0.2965$ & $0.0639$ & $0.3298$ & $0.2493$ & $0.3975$ &$1.20 \times 10^{-2}$ & 0.99\\ 
      Swish & $0.2035$ & $0.1649$ & $0.2838$ & $0.3156$ & $0.3010$ & $0.0590$ & $0.3102$ & $0.2523$ & $0.3681$ &$1.44 \times 10^{-2}$ & 1.11\\
      Softmax & $0.3638$ & $-0.5830$ & $0.5642$ & $0.7392$ & $0.7715$ & $-0.2633$ & $0.8275$ & $0.5142$ & $0.9088$ &$5.87 \times 10^{-3}$ & 1.55\\
      GELU & $0.1714$ & $0.1384$ & $0.2334$ & $0.2790$ & $0.2493$ & $0.0600$ & $0.2584$ & $0.2439$ & $0.2998$ &$6.35 \times 10^{-3}$ & 1.45\\
      Mish & $0.1847$ & $0.1541$ & $0.2618$ & $0.3005$ & $0.2769$ & $0.0557$ & $0.2964$ & $0.2361$ & $0.3402$ &$1.23 \times 10^{-2}$ & 1.31\\
      \bottomrule
    \end{tabular}
  \end{center}
\end{table}

\section{Conclusions}\label{sec:conc}
We present an efficient PINNs framework called Adaptive Interface-PINNs (AdaI-PINNs) for modeling elliptic interface problems with discontinuous coefficients. AdaI-PINNs employ domain decomposition across interfaces, assigning distinct neural networks to subdomains separated by sharp interfaces. Notably, these neural networks vary only in the slopes of their activation functions, with all other parameters remaining the same. In contrast to the Interface PINNs (I-PINNs) framework, which necessitates predefined activation functions for each subdomain, the proposed approach allows the use of a single activation function with an adaptively trainable slope for each subdomain. This adaptive feature fully automates the AdaI-PINNs framework without requiring preset activation functions like I-PINNs and makes it advantageous for problems where several interfaces coexist. Using 1D, 2D and 3D examples, we demonstrate that AdaI-PINNs achieve better accuracy than I-PINNs at nearly half the computational cost with identical hyperparameters. Looking ahead, extending this methodology to transient, coupled-field, and moving interface problems is of significant interest and will be the subject of our future research.

\section*{Acknowledgments}
\sloppy
Chandrasekhar Annavarapu gratefully acknowledges the support from ExxonMobil Corporation to the Subsurface Mechanics and Geo-Energy Laboratory under the grant SP22230020CEEXXU008957. The support from the Ministry of Education, Government of India and IIT Madras under the grant SB20210856CEMHRD008957 is also gratefully acknowledged. Pratanu Roy’s contribution was performed under the auspices of the U.S. Department of Energy by Lawrence Livermore National Laboratory under Contract DE-AC52-07NA27344. 
\bibliographystyle{elsarticle-num}
\bibliography{bibtex}
\end{document}